%% file: main.tex
\documentclass[journal]{IEEEtran}
\IEEEoverridecommandlockouts
\usepackage{times}  
\usepackage{helvet}  
\usepackage{courier}  
\usepackage[hyphens]{url}  
\usepackage{graphicx} 
\usepackage{caption} 
\usepackage{bibentry}
\usepackage{algorithmic}
\usepackage{amssymb,amsfonts}
\usepackage{amsmath}
 \usepackage{graphicx}
\usepackage{subfigure}
\usepackage{textcomp}
\usepackage{xcolor}
\usepackage{tabularx}
\usepackage{multirow}
\usepackage{booktabs}
\usepackage{caption}
\usepackage{lscape}
\usepackage{array}
\usepackage{breakurl}
\usepackage[vietnamese]{babel}
\usepackage{cite}
\usepackage{amsmath}
\usepackage{amssymb,amsfonts}
\usepackage{algorithmic}
\usepackage{textcomp}
\usepackage{enumerate}

\newcolumntype{Y}{>{\centering\arraybackslash}X}

\usepackage{bbding}

\begin{document}
%
\title{DVFL: A Vertical Federated Learning Method for Dynamic Data}
%
%
%


\author{Yuzhi~Liang*, Yixiang Chen*
\IEEEcompsocitemizethanks{\IEEEcompsocthanksitem Yuzhi Liang and Yixiang Chen are with ICNLab, Peking Univerisity Shenzhen Graduate School. *The two authors contribute equally.  Yuzhi Liang is the corresponding author, email: liangyz@pkusz.edu.cn}
}

\maketitle

\begin{abstract}
Federated learning, which solves the problem of data island by connecting multiple computational devices into a decentralized system, has become a promising paradigm for privacy-preserving machine learning. This paper studies vertical federated learning (VFL), which tackles the scenarios where collaborating organizations share the same set of users but disjoint features. Contemporary VFL methods are mainly used in static scenarios where the active party and the passive party have all the data from the beginning and will not change. However, the data in real life often changes dynamically. To alleviate this problem, we propose a new vertical federation learning method, DVFL, which adapts to dynamic data distribution changes through knowledge distillation. In DVFL, most of the computations are held locally to improve data security and model efficiency. Our extensive experimental results show that DVFL can not only obtain results close to existing VFL methods in static scenes, but also adapt to changes in data distribution in dynamic scenarios.
\end{abstract}

\begin{IEEEkeywords}
Vertical Federated Learning, dynamic
\end{IEEEkeywords}

%
\IEEEpeerreviewmaketitle

\input{Introduction}
\input{RelatedWork}
\input{ProbStat}

\input{Experiment2}
\section{Conclusion}
This paper proposes Dynamic Vertical Federated Learning (DVFL), a vertical federated learning method for dynamic data. Specifically, we use feature representation estimation and correction to enhance the data representation in the active party and then train a classifier on the active party for classification. DVFL is applicable for both dynamic scenarios and static scenarios. In a dynamic scenario, the data of the passive party increases dynamically, and the distribution of the data arriving at each timestamp may be different. The experimental results show that in the different distribution changes of dynamic data, DVFL is significantly better than fine-tuning and retrain in most cases. The performance of DVFL is slightly worse than joint training, but joint training is much slower than DVFL. A static scenario can be regarded as a special case of a dynamic scenario: all the data obtained by party B from the beginning. Experimental results show that the performance of DVFL in the static scenario is also competitive with baseline methods.








\bibliographystyle{IEEEtran}
\bibliography{ref}

\end{document}

%% file: Introduction.tex
\section{Introduction}


As an emerging machine learning paradigm, federated learning (FL) enables data owners to collaboratively train models by sharing gradients instead of raw data. The core idea of federated learning is to let each client perform calculations locally on its data to obtain certain intermediate results (e.g., gradients) and then exchange the results with other clients in a secure manner. Existing FL work mainly focuses on the horizontal setting, including designing different model aggregation algorithms \cite{DBLP:journals/isci/ChenRT18,DBLP:conf/aistats/McMahanMRHA17,DBLP:conf/nips/DinhTN20} or solving data Non-IID issues \cite{DBLP:conf/aaai/HuangCZWLPZ21,DBLP:conf/iclr/LiJZKD21}. 

There are a few studies on vertical federated learning (VFL). In VFL, the participants share the same example ID space but are different in feature space. Existing VFL research mainly implements different machine learning algorithms, such as decision trees  \cite{DBLP:journals/pvldb/WuCXCO20,DBLP:journals/corr/abs-1901-08755} and deep learning \cite{DBLP:journals/corr/abs-2008-10838,DBLP:conf/ijcai/ZhangWWXP18}, in the context of data privacy-preserving. Nonetheless, the existing VFL algorithm has the following problems. \textbf{First}, some existing methods (e.g.,\cite{DBLP:conf/ijcai/ZhangWWXP18}) involve a large amount of data interaction between the active party and the passive party and use homomorphic encryption to encrypt data to ensure data security, which requires a lot of computing resources. \textbf{Second}, the existing VFL only considers static scenarios, that is, the participants in the federated learning have all the data from the beginning, and it does not change. However, in real life, data usually grows dynamically, making the overlapping samples between participants in the VFL continue to increase. Intuitively, machine learning methods designed for static scenarios can be updated by fine-tuning, but fine-tuning can only work under the assumption that the distribution of new and old data is similar. This assumption is not always true in real life.  In many scenarios, the distribution of new data is different from that of the original data. In these cases, using fine-tuning to update the model will encounter the problem of catastrophic forgetting. Specifically, when the data distribution of the new data is different from the old data, it means that the model needs to acquire knowledge from the non-stationary data distribution, and the new knowledge will interfere with the old knowledge. Then, using fine-tuning to update the model will cause the model to overwrite or forget the knowledge learned from the old data. 






To alleviate the problems mentioned above, we propose a novel VFL method for dynamic data named \textbf{D}ynamic \textbf{V}ertical \textbf{F}ederated \textbf{L}earning (DVFL for short). Compared to previous methods, the contributions of our work are:


\begin{itemize}
\item DVFL is suitable for dynamic scenarios of vertical federated learning, that is, participants do not acquire all the data at the beginning, the data increases dynamically, and the data distribution of the new data is not necessarily the same as that of the old data.

\item In DVFL, model training is performed locally as much as possible, which can reduce the interaction between parties, thereby improving data security and model efficiency.
\item DVFL does not require participants to share their original data or data encoded by a single neural network.
 \end{itemize}

To evaluate the performance of DVFL in different scenarios, we conducted a lot of experiments on benchmark data sets. The results show that the performance of DVFL in static scenarios is comparable to that of the baseline methods, and it has high efficiency and effectiveness in dynamic scenarios.


%% file: RelatedWork.tex
\section{Related Work}
Vertical Federated Learning (VFL) refers to the technology of federated learning under the setting of different feature spaces for all parties. Different from horizontal federated learning that each client can calculate the loss independently, VFL requires multiple parties to complete the calculation and optimization of the loss function under the framework of security and confidentiality. The existing VFL methods can be divided into linear-based methods, tree-based methods, kernel-based methods, and neural network-based methods. Linear model-based VFL methods include \cite{DBLP:journals/iacr/GasconSB0DZE16, DBLP:journals/corr/abs-1711-10677, DBLP:conf/sp/MohasselZ17}. They use hybrid MPC (secure multi-party computing) protocol \cite{DBLP:conf/focs/Yao82b} or additive homomorphic encryption \cite{DBLP:conf/eurocrypt/BrickellY87} for secure linear model training. Tree-based VFL models include \cite{DBLP:journals/pvldb/WuCXCO20,DBLP:journals/corr/abs-1901-08755}. They enable participating parties to collaboratively build a tree or an forest without information leakage by designing sepecial protocols. Kernel-based VFL methods include \cite{DBLP:series/lncs/DangGH20, DBLP:conf/kdd/GuDLH20}, they approximate the kernel function and federatedly updated the prediction function by the designed gradient. Neural network-based methods include \cite{DBLP:journals/corr/abs-2008-10838,DBLP:conf/ijcai/ZhangWWXP18}. These methods use the active and passive parties to calculate the loss to optimize parameters. Homomorphic encryption is often used to ensure information security.

%% file: ProbStat.tex
\section{Problem Statement}
We consider the problem of dynamic vertical federated learning. Let $\mathcal{D} = \{\textbf{x}_i\}_{i=1}^N$ be the dataset distributed on different parties and the examples are aligned by using encrypted entity alignment techniques \cite{DBLP:journals/corr/abs-1803-04035}. The active party A holds a dataset $\mathcal{D}^A = \{\textbf{x}_i^A\}_{i=1}^N$ and the label $Y = \{y_i\}_{i=1}^N$, where $y_i \in \{0,1\}^C$, $C$ is the number of classes. The passive party B holds a dataset whose size increases over time. At timestamp $t$, party B holds dataset $\mathcal{D}_t^{B} = \{\textbf{x}_i^{B}\}_{i=1}^{N_t^{B}}$, where $t \in \{0,1,...,T\}$ and $N_T^{B} = N$. The increased data of party B from $t-1$ to $t$ is $\bigtriangleup \mathcal{D}_t^B = \mathcal{D}_t^B - \mathcal{D}_{t-1}^B$. Our goal is to design an algorithm that satisfies the following restrictions.

\begin{enumerate}
    \item $\mathcal{D}^A$ and $\mathcal{D}_t^{B}$ cannot be exposed to each other. 
    \item $\mathcal{D}^A$ uses the data of $\mathcal{D}_t^B$ under the privacy protection setting to help improve the performance of the classification model. 
    \item The proposed algorithm should be able to adapt to the dynamic changes of the passive dataset. At each timestamp $t$, even if the data distribution in $\bigtriangleup \mathcal{D}_t^{B}$ is different from that in $\mathcal{D}_t^{B}$, the proposed algorithm should adjust its parameters in a computationally efficient way.
\end{enumerate}

%% file: Experiment2.tex
\section{Experimental Setup}
\subsection{Dataset}
We choose 4 benchmark datasets used in previous studies. 
\begin{itemize}
    \item \textbf{Breast Cancer Wisconsin (BCW)} \cite{bennett1992robust}: The features describe characteristics of the cell nuclei present in the image of a fine needle aspirate (FNA) of a breast mass. It is worth noting that BCW is an imbalanced dataset, and the ratio of positive class to negative class is around 2:8. The dataset is available at \url{https://archive.ics.uci.edu/ml/datasets/Breast+Cancer+Wisconsin}. 
    \item \textbf{Default of Credit Card Clients (DCC)} \cite{DBLP:journals/eswa/YehL09a}: This dataset contains information on credit card clients in Taiwan from April 2005 to September 2005. The dataset is available at \url{http://archive.ics.uci.edu/ml/datasets/default+of+credit+card+clients}.
    \item \textbf{Epsilon (ESP)}: EPS is a dataset of mock data, and ESP5k is a modified version of the ESP dataset used in FATE. The dataset is available at \url{https://github.com/FederatedAI/FATE/blob/master/examples/data/README.md\#epsilon\_5k}
    \item \textbf{Human Activity Recognition (HAR)} \cite{DBLP:conf/sensys/StisenBBPKDSJ15}: The database was built from the recordings of 30 study participants performing activities of daily living. The data is available at \url{https://archive.ics.uci.edu/ml/datasets/Human+Activity+Recognition+Using+Smartphones}.
\end{itemize}

The statistics of the datasets are shown in Table \ref{table_data_size}. 
\begin{table}
\small
\centering
\caption{Dataset statistics}
\label{table_data_size}
\begin{tabular}{c|cc|c|c} 
\toprule
\multirow{2}{*}{\textbf{Dataset}} & \multicolumn{2}{c|}{\textbf{Sample}} &
\multirow{2}{*}{\textbf{Feature}} & \multirow{2}{*}{\textbf{Class}}  \\
\cline{2-3}
& \textbf{Train}  & \textbf{Test}&  & \\
\hline
\textbf{BCW}                               & 453    & 114                        & 32                                 & 2                                \\
\textbf{DCC}                               & 24,000 & 8,000                        & 24                                 & 2                                 \\
\textbf{EPS5k}                               & 4,000  & 1,000                       & 100                                & 2                                \\
\textbf{HAR}                               & 8,239  & 2,060                      & 561                                & 6                            \\
\bottomrule
\end{tabular}
\end{table}

\subsection{Parameter setting} 
The parameter setting in our experiment is as follows. The length of the representation in party p $d^p = 200$, $p \in (A, B)$. The encoder in party A $h^A$ is implemented by a one-layer neural network. The number of hidden units in the neural network is 100 for datasets DCC and EPS5k, and 500 for datasets BCW and HAR.  REN is implemented by a 4 layer neural network, in which each layer has 40 hidden units. The perturbing magnitude $\delta$ for dataset DCC, BCW, EPS5K and HAR are 0.6, 1, 0.6, 0.5, respectively. The batch size is set to 128. The learning rate of experiments on DCC, BCW, EPS5k is 0.005, and that of experiments on HAR is 0.001. Temperature scalar $F = 2$, parameter $\lambda = 0.95$.

\section{Results}
We now analyze the results to answer several research questions. Marco-P, Marco-R, and Marco-F1 are used as our evaluation metrics since BCW and DCC are label imbalanced datasets. We use 5-fold cross validation in our experiments. Our experiments are conducted on a machine running Linux with NVIDIA 1080.

\subsection{RQ1: How does DVFL perform compared to other VFL methods in static settings?}

\begin{table*} [ht]
\centering
\caption{Performance comparison of VFL method in static scenarios}

\begin{tabularx}{.85\textwidth}{@{}Y|l|YY|YYY@{}}

\toprule
&    & \multicolumn{2}{c|}{\textbf{Non-Fed}} & {\multirow{2}{*}{\textbf{Hetero-NN}}} & {\multirow{2}{*}{\textbf{Hetero-SBt}}} & {\multirow{2}{*}{\textbf{DVFL}}}  \\
\cline{3-4}
&    & \textbf{without B} & \textbf{with B}           &                       &                        &  \\ 
\hline
\multirow{3}{*}{\textbf{BCW}}   & P  & 0.8479    & 0.9664           & 0.9320                                     & 0.9444                                     & \textbf{0.9484}                                  \\
                       & R  & 0.9369    & 0.9667           & 0.9436                                     & \textbf{0.9563}                                     & 0.9461                                  \\
                       & F1 & 0.8399    & 0.9661           & 0.9372                                     & \textbf{0.9497}                                     & 0.9465                                  \\
                       \hline

\multirow{3}{*}{\textbf{DCC}}   & P  & 0.6938    & 0.7320           & 0.5819                                     & 0.6478                                     & \textbf{0.7151}                                  \\
                       & R  & 0.6477    & 0.6715           & 0.6874                                     & \textbf{0.7572}                            & 0.6719                                           \\
                       & F1 & 0.6624    & 0.6902           & 0.5913                                     & 0.6731                                     & \textbf{0.6859}                                  \\
                       \hline

\multirow{3}{*}{\textbf{EPS5k}} & P  & 0.5523    & 0.6133           & 0.6051                                     & 0.5967                                     & \textbf{0.6085}                                  \\
                       & R  & 0.5523    & 0.6103           & \textbf{0.6160}                            & 0.5967                                     & 0.6052                                           \\
                       & F1 & 0.5521    & 0.6074           & 0.5969                                     & 0.5967                            & \textbf{0.6025}                                           \\
                       \hline
\multirow{3}{*}{\textbf{HAR}}   & P  & 0.6659    & 0.9009           & 0.5015                                     & 0.8722                & \textbf{0.8982}                                  \\
                       & R  & 0.6669    & 0.8989           & 0.5160                                     & 0.8712                 & \textbf{0.8947}                                  \\
                       & F1 & 0.6483    & 0.898            & 0.4293                                     & 0.8715                 & \textbf{0.8936}                                  \\
\bottomrule
\end{tabularx}
\label{table_performance}
\end{table*}

We use the following methods as baselines in static scenarios (i.e., $T$ =0). 

\begin{itemize}
\item \textbf{Hetero-NN} \cite{DBLP:conf/ijcai/ZhangWWXP18}: Hetero-Neural Network is a neural network-based VFL method implemented in FATE\footnote{https://github.com/FederatedAI/FATE}. For each dataset we use, Hetero-NN has corresponding parameter settings in FATE (\url{https://github.com/FederatedAI/FATE/tree/master/examples/benchmark_quality/hetero_nn}). Thus we use these settings in our experiments.

\item \textbf{Hetero-SBt} \cite{DBLP:journals/corr/abs-1901-08755}: Hetero-Secure Boost is a decision tree-based VFL implemented in FATE. For each dataset we used, Hetero-SBt has corresponding parameter settings in FATE (\url{https://github.com/FederatedAI/FATE/tree/master/examples/benchmark_quality/hetero_sbt}). Thus we use these settings in our experiments.

\item \textbf{Non-federated without party B}: This model consists of an auto-encoding module and a classification module. The implementation of the module is the same as the auto encoding module and classification module in DVFL but only uses the data on party A for prediction. The result can be regarded as the lower bound of DVFL.

\item \textbf{Non-federated with party B}: This model consists of an auto-encoding module and a classification module. The implementation of the module is the same as the auto encoding module and classification module in DVFL. But the encoded data of party A and party B is simply concatenated and then be input into the classifier. The result of this model can be (roughly) regarded as the upper bound of DVFL. 
\end{itemize}

The results are displayed in Table \ref{table_performance}. From the table, we have the following observations.

\textbf{First}, compared with other VFL methods, DVFL obtains the best F1 scores on two datasets (i.e., DCC, EPS5k) while Hetero-SB has the best F1 score on DCC. In general, when the prediction task is more complex (e.g., more features or more types of labels), the advantages of DVFL are more significant.

\textbf{Second}, we can notice that Hetero-SBt has the highest recall rate on the label imbalanced datasets DCC and BCW. This is because Hetero-SBt is a tree-based method whose hierarchical structure allows it to learn signals from both classes. However, the precision of the tree-based method is lower than that of the neural network-based approach, which affects the overall F1 score.

\textbf{Third}, the performance of Hetero-NN is not good, partly because it involves many encryption and decryption operations. With limited computing resources, it can only support simple models (such as fewer neural network layers and hidden units), which is insufficient for complex datasets. 

\subsection{RQ2: Does DVFL perform well in dynamic data with different data distributions?}
\begin{table*}
\centering
\small
\caption{Performance comparison of different model update methods in dynamic scenarios}
\label{Tabl:Diff_dist}
\begin{tabular}{c|c|c|cccc} 
\toprule
\multirow{2}{*}{}                               
{\multirow{2}{*}{\textbf{Mode}}}            
& \multicolumn{1}{c|}{\multirow{2}{*}{\textbf{Timestamp}}} & \multirow{2}{*}{\textbf{Class Ratio (Pos:Neg) }} & \multicolumn{4}{c}{\textbf{ macro-F1 }}                          \\ 
\cline{4-7}
                                                                            & \multicolumn{1}{c|}{}                                    &                                                  & Retrain       & Fine-Tune      & DVFL(Ours)     & Joint Training  \\ 
\hline
\multirow{6}{*}{\textbf{Random}}                                                     & 0                                                        & ~16.7\% : 16.7\% (5:5)                           & 0.572         & 0.572          & 0.572          & 0.572           \\
                                                                            & 1                                                        & 23.3\% :10.0\% (7:3)                             & 0.462         & 0.460          & \underline{0.547}  & \textbf{0.586}  \\
                                                                            & 2                                                        & 20.0\% : 13.3\% (6:4)                            & \underline{0.550} & 0.522          & 0.546          & \textbf{0.555}  \\
                                                                            & 3                                                        & ~ 3.3\% : 30.0\% (1:9)                           & 0.343         & 0.347          & \textbf{0.607} & \underline{0.593}   \\
                                                                            & 4                                                        & 13.3\% : 20.0\% (4:6)                            & \underline{0.581} & 0.561          & 0.580          & \textbf{0.602}  \\
                                                                            & 5                                                        & 23.3\% : 10.0\% (7:3)                            & 0.520         & 0.450          & \textbf{0.598} & \underline{0.585}   \\ 
\hline
\multirow{6}{*}{\begin{tabular}[c]{@{}c@{}}\textbf{Asc vs Des}\end{tabular}} & 0                                                        & 20.0\% : 20.0\% (1:1)                            & 0.598         & 0.598          & 0.598          & 0.598           \\
                                                                            & 1                                                        & 28.8\% : 3.2\% (9:1)                             & 0.406         & 0.368          & \underline{0.585}  & \textbf{0.604}  \\
                                                                            & 2                                                        & 22.4\% : 9.6\% (7:3)                             & \underline{0.542} & 0.453          & \textbf{0.561} & 0.540           \\
                                                                            & 3                                                        & 16\% : 16\% (5:5)                                & 0.602         & \textbf{0.607} & 0.584          & \underline{0.603}   \\
                                                                            & 4                                                        & 9.6\% : 22.4\% (3:7)                             & 0.501         & 0.469          & \underline{0.562}  & \textbf{0.615}  \\
                                                                            & 5                                                        & 3.2\% : 28.8\% (1:9)                             & 0.378         & 0.332          & \textbf{0.613} & \underline{0.606}   \\ 
\hline
\multirow{6}{*}{\textbf{Parallel}}                                                    & 0                                                        & 50\% : 20\% (5:2)                                & 0.545         & 0.545          & 0.545          & 0.545           \\
                                                                            & 1                                                        & 10\% : 16\% (5:8)                                & 0.409         & 0.524          & \textbf{0.590} & \underline{0.564}   \\
                                                                            & 2                                                        & 10\% : 16\% (5:8)                                & 0.562         & 0.547          & \underline{0.609}  & \textbf{0.612}  \\
                                                                            & 3                                                        & 10\% : 16\% (5:8)                                & 0.336         & 0.543          & \underline{0.611}  & \textbf{0.634}  \\
                                                                            & 4                                                        & 10\% : 16\% (5:8)                                & 0.541         & 0.508          & \textbf{0.629} & \underline{0.573}   \\
                                                                            & 5                                                        & 10\% : 16\% (5:8)                                & 0.572         & 0.558          & \underline{0.605}  & \textbf{0.616}  \\ 
\hline
\multirow{6}{*}{\textbf{Uniform}}                                                       & 0                                                        & 25\% : 25\% (1:1)                                & 0.599         & 0.599          & 0.599          & 0.599           \\
                                                                            & 1                                                        & 15\% : 15\% (1:1)                                & 0.579         & \textbf{0.602} & 0.580          & \underline{0.586}   \\
                                                                            & 2                                                        & 15\% : 15\% (1:1)                                & 0.575         & \underline{0.609}  & 0.590          & \textbf{0.610}  \\
                                                                            & 3                                                        & 15\% : 15\% (1:1)                                & 0.574         & \textbf{0.614} & 0.587          & \underline{0.602}   \\
                                                                            & 4                                                        & 15\% : 15\% (1:1)                                & 0.580         & \textbf{0.612} & 0.580          & \underline{0.607}   \\
                                                                            & 5                                                        & 15\% : 15\% (1:1)                                & 0.584         & \textbf{0.619} & 0.587          & \underline{0.603}   \\
\bottomrule
\end{tabular}
\end{table*}


We evaluated the performance of DVFL under differently distributed data streams on the EPS5k dataset. EPS5k is a dataset for a binary classification task. In our experiment, we assume that the data of party B arrives in $T$ times. At timestamp $t$, party B obtains the dataset $\bigtriangleup \mathcal{D}_{t}^B = \mathcal{D}_{t}^B - \mathcal{D}_{t-1}^B$. Our task is to use $\bigtriangleup \mathcal{D}_{t}^B = \mathcal{D}_{t}^B - \mathcal{D}_{t-1}^B$ and the corresponding data in party A to train the classifier of DVFL in a privacy-preserving manner. 

To measure the performance of DVFL under the different distribution of data streams, we use the following 4 modes of data distributions:
\begin{itemize}
\item \textbf{Random}: For each timestamp, the ratio of positive and negative examples in the new data is random.
\item \textbf{Asc vs Des}: Over time, the number of positive examples in the new data gradually increases, while the number of negative examples gradually decreases.
\item \textbf{Parallel}: At each timestamp, the ratio of positive and negative examples in the new data is the same but imbalanced.

\item \textbf{Uniform}: At each timestamp, the ratio of positive and negative examples in the new data is 1:1. 
\end{itemize}
Finally, we use the trained classifier to classify the items in the test set and record the results. In the test set, the ratio of positive and negative examples is roughly 1:1. 
We use the following model update methods as our baselines:
\begin{itemize}
\item \textbf{Fine-tuning}: Fine-tuning uses the new dataset to tune the current classifier with a small learning rate (0.1 times the original learning rate); 
\item \textbf{Joint Training}: Using All uses all previously shown data to train a new classifier, which should be the highest possible result in most cases.
\item \textbf{Retrain}: Retrain uses only the newly arrived dataset to train a new classifier.
\end{itemize}
The results are in Table ~\ref{Tabl:Diff_dist}. From the table, we have the following observations.



\textbf{First}, in different modes, the performance of DVFL is much better than retrain and fine-tuning, especially when the data distribution of the new data is very different from that of the old data. This means that DVFL has better adaptability to changes in the data distribution of dynamic data. 

\textbf{Second}, in the mode where the data distribution is relatively stable (e.g., parallel, uniform), DVFL also has good performance. However, the performance difference between different methods in these modes is small, especially in the uniform mode. In theory, the performance of fine-tuning, joint training, and DVFL in uniform mode is basically the same.

\textbf{Third}, when the data distribution of training data and test data is similar, the performance of joint training is the best. However, the training time required for joint training is much longer than of other methods. This is because that at each timestamp, joint training works on the entire dataset, while the rest of the methods only works on the new data. It is worth noticing that the performance of joint training in Table \ref{Tabl:Diff_dist} is not the best in all cases. This is because the ratio of positive and negative examples in the test set is close to 1:1, but since the data in party B is dynamically increasing, there is a difference in the label distribution of the test data and the training data at some specific timestamps. When $t = T$, party B obtains all the data.

\begin{figure}[ht]
	\centering  
	\subfigure[BCW]{
		\includegraphics[width=0.45\linewidth]{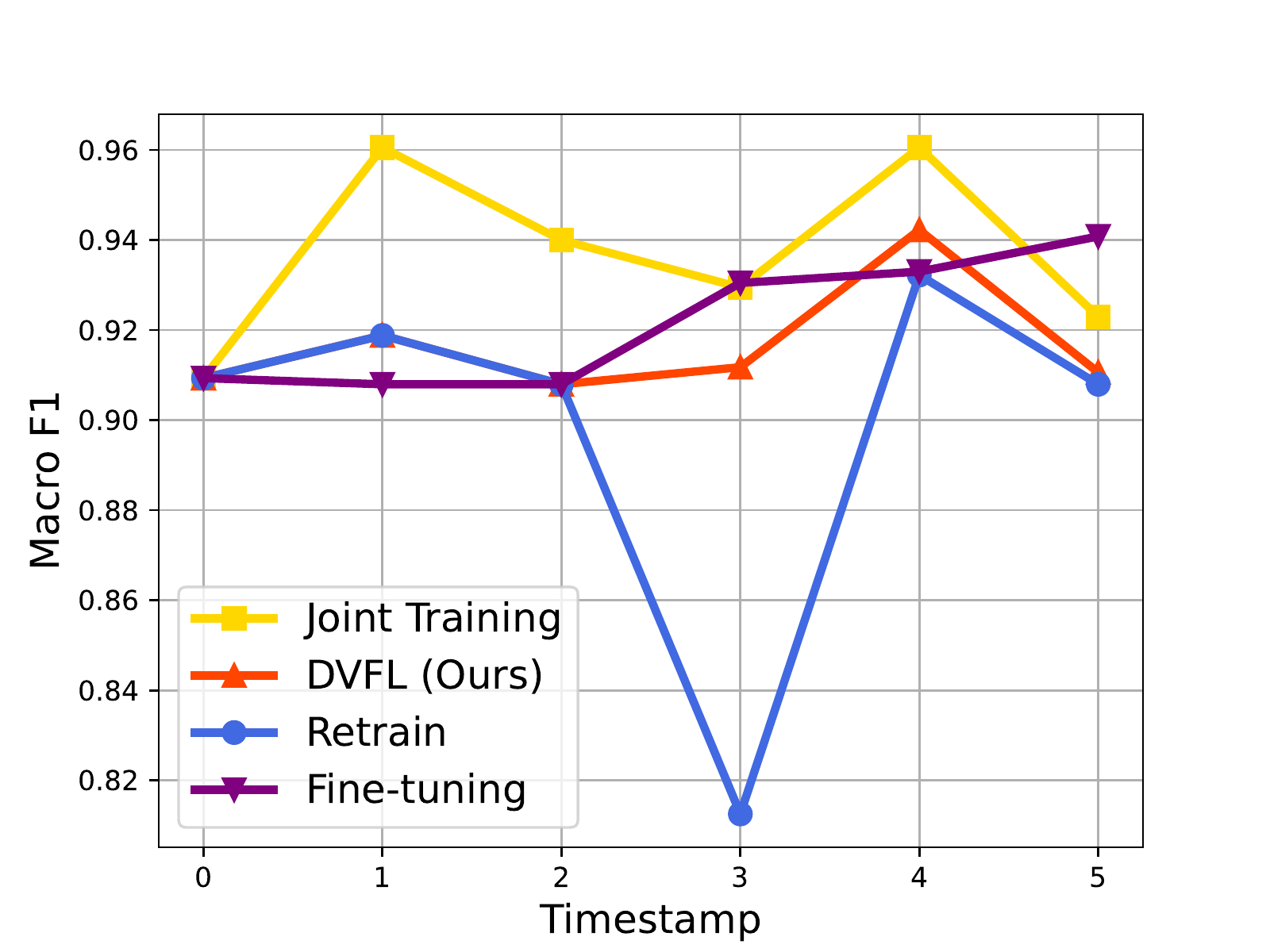}}
	\subfigure[DCC]{
		\includegraphics[width=0.45\linewidth]{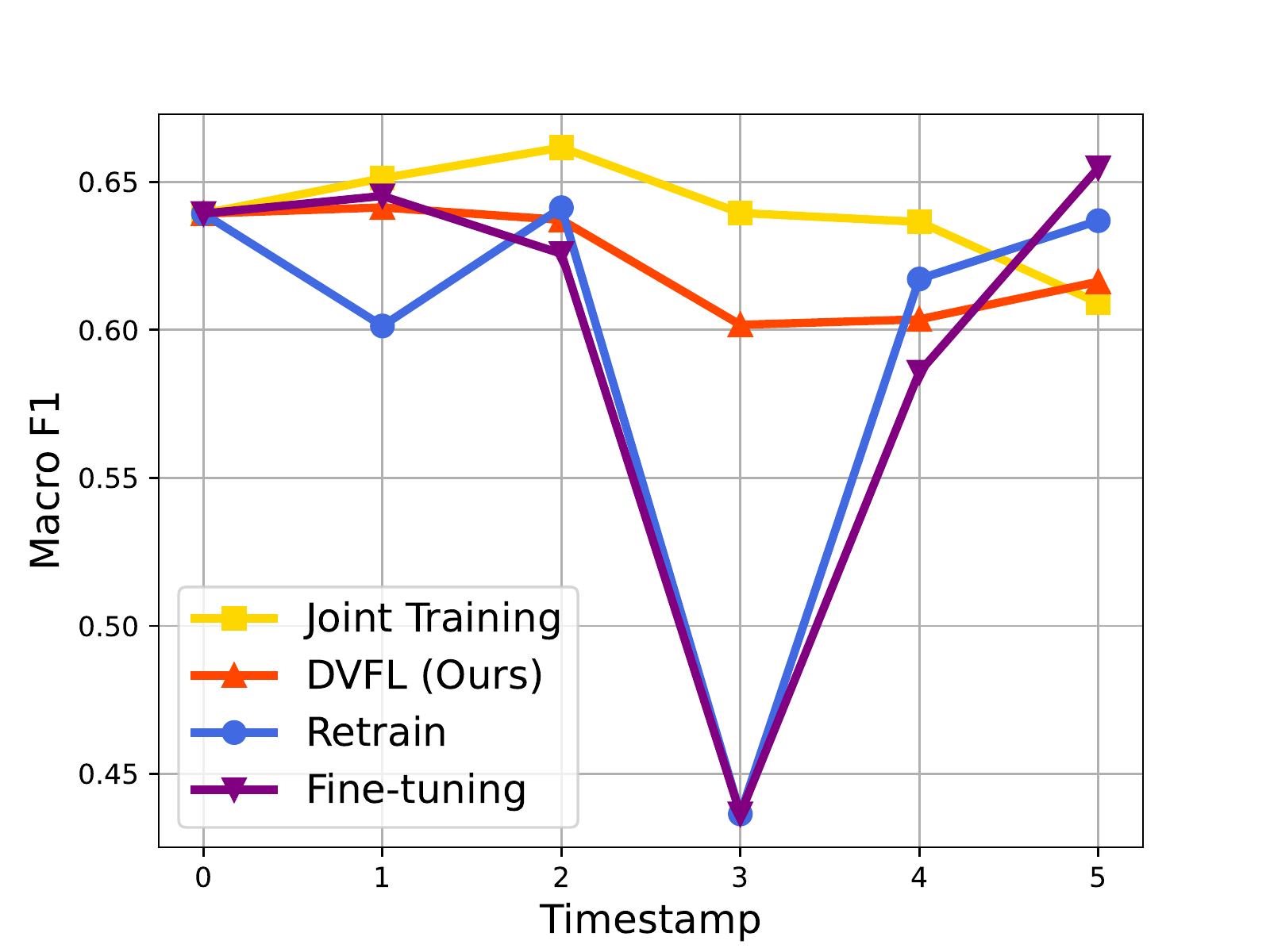}}
	\caption{Comparison of different model update methods on DCC and BCW (random mode)}
\label{Fig_otherset}
\end{figure}



To further evaluate the performance of DVFL on other datasets, we tested its results of random mode on BCW and DCC. As Fig. \ref{Fig_otherset}, the results on these two datasets are consistent with the results on EPS5k.

\subsection{RQ3: How does DVFL perform when the number of clients in the passive party increases?}

\input{table_multi_passive}


\begin{table}
\centering
\caption{Scalability evaluation of DVFL}
\label{Tab:Multi_passive}
\begin{tabular}{c|ccc} 
\toprule
\textbf{Passive client \#} & \textbf{P} & \textbf{R} & \textbf{F1}  \\
\hline
1 & 0.6300    & 0.6698 & 0.6280    \\
2 & 0.5927    & 0.7382 & 0.5921    \\
4 & 0.6036    & 0.7138 & 0.6084    \\
6 & 0.5964    & 0.6934 & 0.5964    \\
8 & 0.6078    & 0.6629 & 0.5839    \\
10& 0.5803    & 0.7637 & 0.5821    \\
\bottomrule
\end{tabular}
\end{table}

To evaluate the scalability of DVFL, we tested the performance when the passive party has multiple clients. Specifically, we measure the scalability of DVFL on the EPS5K dataset, and the number of passive clients $q$ ranges from 1 to 10. Each client on the passive side has $\frac{1}{q}$ of $\mathcal{D}^B$. The results are demonstrated in Table \ref{Tab:Multi_passive}. It can be seen from the table that when the number of clients on the passive party increases, the performance of the system is still stable.